\documentclass[10pt,twocolumn,letterpaper]{article}

\usepackage{cvpr}              %

\usepackage[dvipsnames]{xcolor}
\usepackage{comment}
\usepackage{dsfont}
\usepackage{multirow}
\usepackage[accsupp]{axessibility}

\definecolor{cvprblue}{rgb}{0.21,0.49,0.74}
\usepackage[pagebackref,breaklinks,colorlinks,allcolors=cvprblue]{hyperref}

\usepackage[capitalize]{cleveref}

\newcommand{\OURS}{LT3SD}
\newcommand{\myparagraph}[1]{\vspace{4pt}\noindent\textbf{#1}}

\DeclareMathOperator*{\argmin}{arg\,min}

\newcommand{\Indicator}{\mathds{1}}

\definecolor{lightblue}{HTML}{9BD2FF}
\definecolor{coralpink}{HTML}{EC7C87}

\crefname{section}{Sec.}{Secs.}
\Crefname{section}{Section}{Sections}
\Crefname{table}{Table}{Tables}
\crefname{table}{Tab.}{Tabs.}

\def\paperID{6520} %
\def\confName{CVPR}
\def\confYear{2025}

\title{LT3SD: Latent Trees for 3D Scene Diffusion}

\author{Quan Meng \hspace{1cm} Lei Li \hspace{1cm} Matthias Nie{\ss}ner \hspace{1cm} Angela Dai \vspace{0.2cm} \\
Technical University of Munich\vspace{0.2cm}\\
\href{https://quan-meng.github.io/projects/lt3sd/}{quan-meng.github.io/projects/lt3sd}
\vspace{-0.4cm}
}

\begin{document}
\twocolumn[{
    \renewcommand\twocolumn[1][]{#1}
    \maketitle
    \begin{center}
    \vspace{-16pt}
    \centering
        \includegraphics[width=1.0\linewidth]{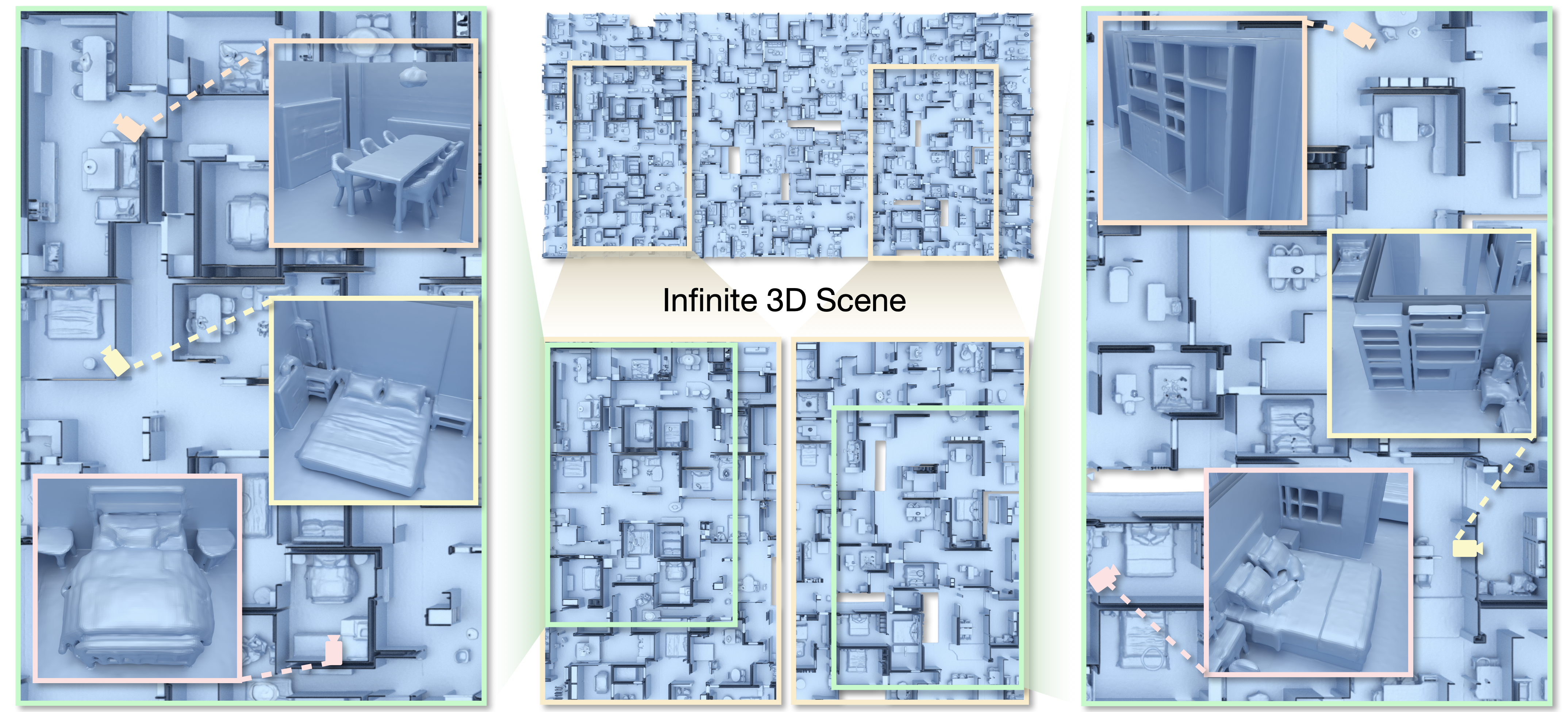}
    \vspace{-0.7cm}
    \captionof{figure}{We introduce \OURS{}, a novel latent 3D scene diffusion approach enabling high-fidelity generation of infinite 3D environments. We train \OURS{} on a latent tree-based 3D scene representation, encoding both lower-frequency geometry and higher-frequency detail, and synthesize infinite scenes in a patch-by-patch and coarse-to-fine fashion. }
    \label{fig:teaser}
    \end{center}
}]

\begin{abstract}

We present \OURS{}, a novel latent diffusion model for large-scale 3D scene generation. Recent advances in diffusion models have shown impressive results in 3D object generation, but are limited in spatial extent and quality when extended to 3D scenes. To generate complex and diverse 3D scene structures, we introduce a latent tree representation to effectively encode both lower-frequency geometry and higher-frequency detail in a coarse-to-fine hierarchy. We can then learn a generative diffusion process in this latent 3D scene space, modeling the latent components of a scene at each resolution level. To synthesize large-scale scenes with varying sizes, we train our diffusion model on scene patches and synthesize arbitrary-sized output 3D scenes through shared diffusion generation across multiple scene patches. Through extensive experiments, we demonstrate the efficacy and benefits of \OURS{} for large-scale, high-quality unconditional 3D scene generation and for probabilistic completion for partial scene observations.

\vspace{-0.4cm}

\end{abstract}
    
\section{Introduction}
\label{sec:intro}

In recent years, there has been increasing demand for 3D digital content, both for content creation purposes such as films, video games, and mixed reality, as well as for reconstruction to enable machines to perceive real environments from visual inputs.
For instance, video games have driven efforts to produce realistic virtual 3D environments, requiring significant costs for highly-trained 3D artists to create a vast array of 3D scene assets. In particular, AAA video game budgets have risen dramatically in the last five years, with games planned to be released this year and next having development budgets of \$200 million or more \cite{ukcma}.

Recent advances in diffusion models~\cite{sohl2015deep, ho2020denoising, rombach2021highresolution, 2024SoraReview, peebles2023scalable, ramesh2021zero} have achieved remarkable success in generating complex, high resolution, realistic images, and videos.
In the 3D domain, however, generative modeling remains a significant challenge due to the data and representation complexity and limited training data availability.
In addition, signal in 3D content is highly unevenly distributed, with many regions being predominantly free space, whereas most scene detail is located only around certain surface areas.
Existing 3D diffusion models~\cite{chou2022diffusionsdf,erkocc2023hyperdiffusion,li2023diffusionsdf,luo2021diffusion,nam20223d,shue20233d,ssdnerf, zeng2022lion} overwhelmingly focus on object-level generation. Following latent diffusion (LDMs)~\cite{rombach2021highresolution} for 2D images, current state-of-the-art methods focus on training diffusion models on learned implicit representations of 3D shapes. However, these methods assume that shapes lie in a canonical and bounded space that is amenable to fixed and compact representation encodings, such as global latent codes \cite{chou2022diffusionsdf}, hypernetworks \cite{erkocc2023hyperdiffusion}, and triplanes \cite{shue20233d,gupta20233dgen}. While these representations yield impressive results for class-specific distributions, such as chairs or cars, they face challenges when applied to 3D scene environments characterized by highly unstructured geometries, diverse object arrangements, and varying spatial extents. Although early attempts at scene-level synthesis exist, they predominantly focus on single rooms or limited spatial extents~\cite{hollein2023text2room, kim2023nfldm, ren2023xcube} with relatively low-resolution~\cite{lee2023diffusion,liu2023pyramid,lee2024semcity} 3D scenes.

\begin{figure}[t]
	\begin{center}
		\includegraphics[width=1\linewidth]{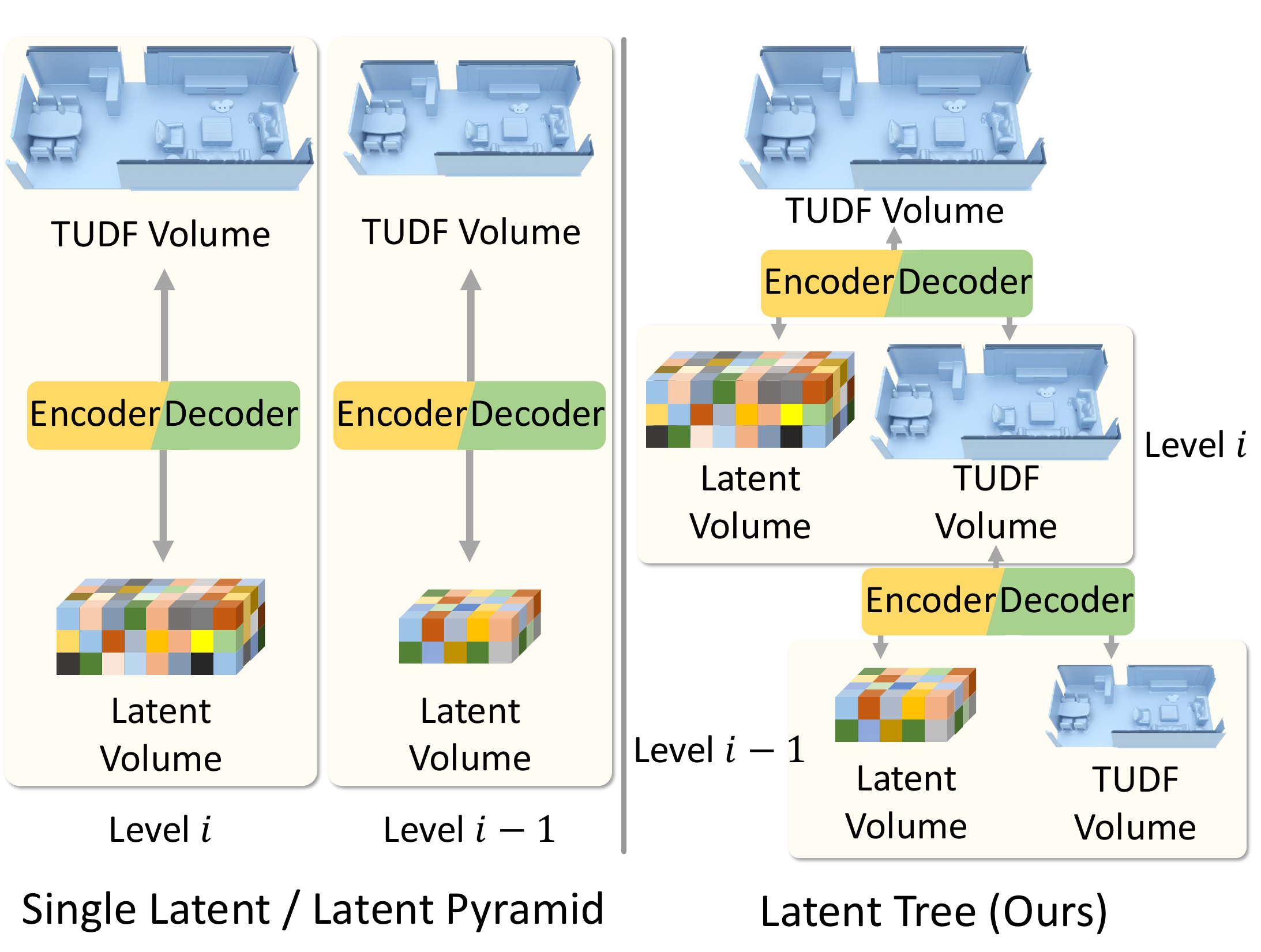}
	\end{center}
	\vspace{-0.25in}
	\caption{\textbf{Latent representations for 3D scenes.} 
     In contrast to encoding a 3D scene into a single latent feature grid or a multi-level latent pyramid on the left, 
     our latent tree representation on the right is a learned hierarchical decomposition with a series of geometry (lower-frequency) and latent feature (higher-frequency) encodings at each resolution level.
        }
    \vspace{-0.3cm}
	\label{fig:latent_tree_pipeline}
\end{figure}

To this end, we propose \OURS{}, a new probabilistic model for generating high-quality, large-scale, arbitrary-sized 3D scenes. Key to our approach is a novel latent tree representation (\cref{fig:latent_tree_pipeline}-right) that progressively decomposes a 3D scene into a tree hierarchy from fine to coarse resolutions. 
Our latent tree representation efficiently factorizes 3D scenes into geometry (lower-frequency) and latent feature (higher-frequency) volumes, enabling a more compact, effective representation for latent scene diffusion.
We then synthesize 3D scenes in a coarse-to-fine, patch-by-patch fashion by reversing the latent tree decomposition.

We introduce a latent tree-based diffusion model to generate the latent feature volume conditioned on the corresponding geometry volume at the same tree level. 
This conditional learning of scene details helps to reduce the complexity of modeling 3D scene distributions.
As 3D scenes can vary significantly in size, we design our diffusion model to learn at the level of scene patches. 
This strategy helps to shift the learning focus from complex unaligned 3D scenes to local structures with higher shared similarity.

As a result, our approach can generate large-scale, even infinite 3D scenes by coarse-to-fine construction of the latent trees in a patch-wise fashion.
Starting from the coarsest level, our method operates patch-wise to generate the overall structure of a 3D scene, which is critical for geometric coherency. Then, in a coarse-to-fine manner, our method conditionally generates the latent feature volume, which represents fine details for each upper level—the final 3D scene results from the inverse process of latent tree decomposition.
Experiments show that our method significantly outperforms existing baselines, improving FID scores by 70\% on this challenging 3D scene generation task.

In summary, we present the following contributions:
\begin{itemize}
    \item We introduce a novel latent 3D scene diffusion approach,
    leveraging a latent tree representation to compactly encode complex 3D scene geometry. This enables high-fidelity 3D scene generation in a coarse-to-fine fashion along the latent tree resolution levels.
    
    \item We characterize 3D scene generation in a patch-wise fashion, enabling efficient training and seamless infinite 3D scene generation through shared diffusion generation across multiple scene patches.
    
\end{itemize}

\section{Related Work}
\label{sec:related_works}

\myparagraph{Neural 3D Scene Representations.}
Various neural representations have been developed to characterize 3D shapes or scenes, with recent works extensively exploring implicit functions~\cite{park2019deepsdf, onet, chen2018implicit_decoder, mildenhall2020nerf}, feature volumes~\cite{Peng2020ECCV, liu2020neural,jiang2020local}, triplanes~\cite{Chan2022,shue20233d}, and tensor decompositions~\cite{Chen2022ECCV}. These representations are proposed as alternatives to traditional representations, such as voxels~\cite{wu20153d,dai2017shape,dai2020sg,dai2021spsg,siddiqui2021retrievalfuse} or point clouds~\cite{yang2019pointflow,achlioptas2018learning,luo2021diffusion}, to encode geometry  more compactly and while maintaining high resolutions. Earlier approaches model an implicit function as an MLP, though such a global representation can result in difficulty retaining both local and global structures, with limited editability~\cite{park2019deepsdf, onet, mildenhall2020nerf}. Tri-planes and tensor decompositions more effectively balance memory cost and performance, making them popular choices for object-level 3D shape generation~\cite{shue20233d, gupta20233dgen}; however, scene outpainting is not straightforward as the $n$-planes are highly correlated, requiring intricate synchronization processing for extrapolation~\cite{wu2024blockfusion}. In this work, we use a feature grid representation for capability to handle arbitrary topologies and its effective feature spatial locality to enable patch-based compositions of large-scale scenes.

\begin{figure*}[t]
	\begin{center}
		\includegraphics[width=1.0\linewidth]{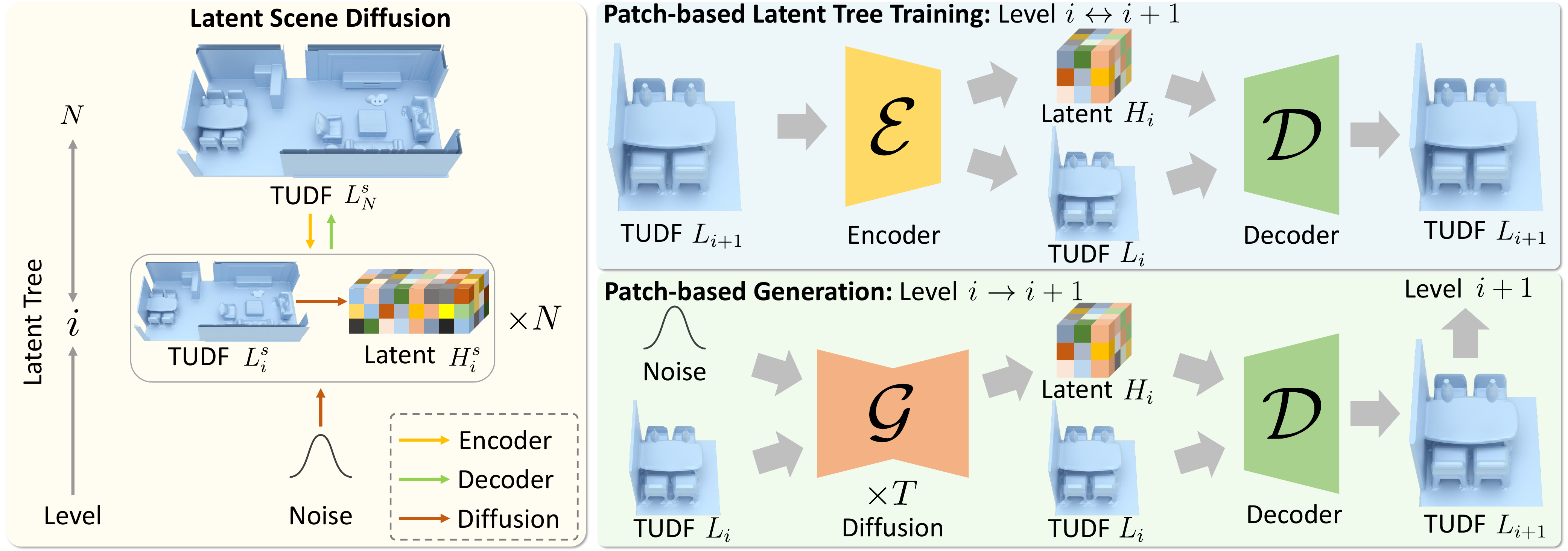}
	\end{center}
	\vspace{-0.5cm}
	\caption{\textbf{Overview of \OURS{}.}
    We formulate 3D scene generation as a patch-based latent diffusion process. 
    \textbf{Left}: To characterize complex scene geometry, we encode 3D scenes in a novel latent tree representation, where each scene resolution level $i \in [1, N-1]$ is decomposed into a TUDF grid $L_{i}^s$ and a latent feature grid $H_{i}^s$. \textbf{Top Right}: During latent tree training, the encoder $\mathcal{E}$ encodes a patch $L_{i+1}$ from the scene grid $L_{i+1}^s$ at resolution level $i+1$ to a coarser TUDF patch $L_i$ and a  latent feature patch $H_i$ at level $i$. The decoder $\mathcal{D}$ then reconstructs the scene patch $L_{i+1}$ based on the factorized grids $L_i$ and $H_i$. 
    \textbf{Bottom Right}:  During generation, the diffusion model $\mathcal{G}$ learns to generate a latent feature patch $H_i$ conditioned on a TUDF patch $L_i$ within the same level $i$. 
    Our method enables arbitrary-sized 3D scene generation at inference time by synthesizing scenes in a coarse-to-fine hierarchy and a patch-by-patch fashion.}
 \vspace{-0.4cm}
\label{fig:pipeline}
\end{figure*}

\myparagraph{3D Diffusion Models for Shape Generation.}
Diffusion models have shown unprecedented capability in generating 2D images and videos~\cite{rombach2021highresolution, blattmann2023align}.
Existing 3D diffusion work has focused on synthesizing 3D shapes  \cite{ShapeGF, erkocc2023hyperdiffusion, shue20233d, zeng2022lion, chou2022diffusionsdf, li2023diffusionsdf, cheng2023sdfusion, gupta20233dgen, ssdnerf, wang2023rodin, xiang2024structured, lan2024ln3diff, chen2024primx}, showing quite promising results. Following LDMs~\cite{rombach2021highresolution}, most 3D diffusion models follow a two-stage approach: first learn a latent space of 3D shapes, and then train diffusion in the latent space. However, extending the same success to complex 3D scenes remains a challenge. Such 3D scene modeling faces higher data dimensionality, but with orders of magnitude less training data. 
Existing object generation methods assume that shapes lie in a canonical space and do not consider the arbitrary-sized nature of 3D scenes, making extending to 3D scenes difficult. In contrast, our approach enables efficient infinite 3D scene generation via training on latent tree scene patches.

\myparagraph{3D Scene Generation.}
Inspired by the success of 2D image generation, various recent methods have proposed reconstructing 3D scenes using 2D generative models~\cite{chan2023genvs, chen2023sd}. These approaches leverage 2D image generative models to learn to generate consistent novel views; however, designing the camera trajectory and generating a complete 3D scene is challenging due to occlusions. Scene graphs-based 3D generation methods~\cite{tang2024diffuscene,zhai2023commonscenes} offer an abstract representation for 3D scene generation. Several recent works have introduced diffusion models for 3D scenes~\cite{ju2023diffroom, lee2023diffusion, Bahmani_2023_ICCV, liu2023pyramid, ren2023xcube, xiang2024structured, wu2024blockfusion, lee2024semcity, kim2023nfldm}. DiffRoom~\cite{ju2023diffroom}, NeuralField-LDM~\cite{kim2023nfldm}, CC3D~\cite{Bahmani_2023_ICCV}, XCube~\cite{ren2023xcube} and TRELLIS~\cite{xiang2024structured} focus on bounded generation of 3D scenes, while SSG~\cite{lee2023diffusion}, \cite{liu2023pyramid}, and SemCity~\cite{lee2024semcity} focus on outdoor 3D semantic scene generation, they offer limited geometry detail due to low-resolution data or explicit representation. 
Text-to-3D-based methods~\cite{hollein2023text2room} typically require hours of optimization. Very recently, BlockFusion~\cite{wu2024blockfusion} proposed to generate larger-scale 3D scenes through latent triplane extrapolation, conditioned on 2D layout inputs. However, 3D scene generation is still in its infancy, and it remains a challenge to efficiently generate unbounded 3D scenes with high-quality, coherent structures, high-frequency details, and plausible layouts. In this work, we propose an efficient unconditional 3D scene generation pipeline that enables seamless infinite 3D scene generation at high quality.

\section{Method}
\label{sec:methods}
\subsection{Overview}
Our goal is to develop a probabilistic model for generating 3D scenes encompassing complex geometric structures, diverse object arrangements, and varying spatial extents.
We propose \OURS{}, a novel coarse-to-fine, patch-by-patch 3D scene generation approach by leveraging diffusion models in the space of a learned latent tree representation for 3D scenes.
Our method has two main stages: latent tree encoding and patch-based latent diffusion, as illustrated in \cref{fig:pipeline}.

In the first stage (\cref{sec:first_stage}), we progressively decompose a 3D scene into a hierarchical latent tree from fine to coarse levels.
Each tree level is a factorization of the scene into a geometry volume and a latent feature volume at the corresponding resolution level (\cref{fig:pipeline}-left).
The geometry volume, represented as a truncated unsigned distance field (TUDF), captures lower-frequency scene information, while the latent feature volume compactly encodes higher-frequency details.
As shown in \cref{fig:pipeline}-top right, we learn this factorization with a 3D CNN-based encoder $\mathcal{E}$.
The inverse process is built upon a decoder $\mathcal{D}$ that combines the two decomposed volumes to accurately recover the geometry volume at a higher resolution level.

In the second stage (\cref{sec:second_stage}), we train a 3D UNet-based diffusion network on the latent tree encodings.
At each level, we randomly crop fixed-size patches from the scene geometry volumes and predict their corresponding latent feature patches (\cref{fig:pipeline}-bottom right).
At inference time, we can effectively synthesize large-scale 3D scenes by progressively generating latent trees in a patch-by-patch fashion from coarse-to-fine levels.

\subsection{Latent Tree Representation for 3D Scenes}
\label{sec:first_stage}

To enable high-quality 3D scene generation, we propose to transform a 3D scene into a latent tree representation.
Each level of the hierarchy comprises factorized geometry and latent feature encodings of the scene at the corresponding resolution.
We represent the geometry encoding as truncated unsigned distance field (TUDF) grids, characterizing lower-frequency scene information, while higher-frequency scene details are encoded into a latent feature grid.

Our latent tree representation, as depicted in \cref{fig:latent_tree_pipeline}-right, offers the following advantages:
(1) It allows for generating complex scene geometry in a coarse-to-fine fashion, in contrast to a single latent (\cref{fig:latent_tree_pipeline}-left), which does not well-represent shared local structures; 
(2) Compared to a multi-level latent pyramid (\cref{fig:latent_tree_pipeline}-left), our latent tree offers better encoding of 3D scenes with higher reconstruction quality.

\myparagraph{Construction.}
To construct the latent tree representations, we first compute UDFs for 3D scenes. UDFs can easily encode arbitrary 3D scene geometry with various topologies. 
To focus on surface geometry regions and reduce data distribution complexity, we clip distances exceeding a specified threshold $\tau$ to obtain TUDF voxel grids.

Concretely, we decompose the TUDF grid of a 3D scene into a latent tree with $N$ levels, where each scene resolution level $i \in [1, N-1]$ consists of a TUDF grid $L_i^s$ and a latent feature grid $H_i^s$. We denote the root of the tree (\ie{}, the highest resolution) as $L_N^s$.
As illustrated in \cref{fig:pipeline}-top right, we learn this decomposition process with fixed-size patches $L_{i+1}$ randomly cropped from the scene grid $L_{i+1}^s$ for computational efficiency. 
All scene levels can be trained in parallel through patch-based training.

As shown in \cref{fig:pipeline}-top right, at resolution level $i+1$, we use the encoder $\mathcal{E}_{i+1}$ to factorize the TUDF patch $L_{i+1} \in \mathbb{R}^{\scriptscriptstyle 1 \times D_{i+1} \times H_{i+1} \times W_{i+1}}$ into a lower-resolution TUDF patch $L_{i}$ and a latent feature patch $H_{i}$ at the next coarser level $i$:
\begin{equation}
    \label{eq:latent_tree_encoder}
    \mathcal{E}_{i+1}(L_{i+1}) \Rightarrow  [L_{i}, H_{i}],
\end{equation}
where $L_{i} \in \mathbb{R}^{\scriptscriptstyle 1 \times D_{i} \times H_{i} \times W_{i}}$ represents the coarser patch geometry, and $H_{i} \in \mathbb{R}^{\scriptscriptstyle C \times D_{i} \times H_{i} \times W_{i}}$ with a latent feature dimension $C$ encodes higher-frequency details.
In practice, we compute $L_{i}$ by downsampling $L_{i+1}$ with average pooling and predict $H_{i}$ with a 3D CNN.

We then train a decoder network $\mathcal{D}_{i+1}$ to reconstruct the TUDF patch $L_{i+1}$ by combining the factorized grids $L_{i}$ and $H_{i}$ as follows:
\begin{equation}
    \label{eq:latent_tree_decoder}
    \mathcal{D}_{i+1}([L_{i}, H_{i}]) \Rightarrow L_{i+1}.
\end{equation}
The decoder $\mathcal{D}_{i+1}$ is also implemented as a 3D CNN.

\myparagraph{Learning.}
To train the encoders and decoders, our loss measures the $\ell_2$ error between the reconstructed TUDFs and the ground truth TUDFs:
\begin{equation}
\mathcal{L}_{\text{latent}} = \Big(L_{i+1} -\mathcal{D}_{i+1}\big(\mathcal{E}_{i+1}\left(L_{i+1}\right)\big) \Big)^2.
\end{equation}

Given the trained encoders, we can progressively decompose a high-resolution, arbitrary-sized 3D scene into a compact latent tree $\{L_1^s, H_1^s, \cdots, L_{N-1}^s, H_{N-1}^s \}$.

\subsection{Patch-Based Latent Scene Diffusion}
\label{sec:second_stage}

We learn a denoising diffusion probabilistic model $\mathcal{G}_i$ \cite{ho2020denoising} for each resolution level $i$ of the latent tree to generate high-fidelity 3D scenes.
We leverage the explicit factorization in our latent tree (\cref{sec:first_stage}) and train $\mathcal{G}_i$ to generate the latent feature grid $H_i^s$, conditioned on geometry grid $L_i^s$. Since 3D scenes typically consist of local structures with shared similarities, our latent diffusion models are trained on scene patches randomly cropped from scene grids. 

We denote the diffusion model as $\mathcal{G}_i (z_t, t, c)$, denoising a noisy latent patch $z_{t}$ at each time step $t$ based on condition $c$.
At each training step, a patch of Gaussian noise $\epsilon\sim\mathcal{N}(0, 1)$ of the same size and a time step $t \in [1, T]$ are randomly sampled and applied to a latent patch $z$ to obtain a noisy latent patch $z_t$. 
At level $i>1$, we train the diffusion model $\mathcal{G}_i$ such that given a coarse geometry patch $L_i$ as condition, it predicts its corresponding latent feature patch $H_i$, \ie{}, $z = H_i$ and $c=L_i$. 
At level $i=1$, the coarsest level of the latent tree, the diffusion model generates both grids $L_i$ and $H_i$ unconditionally, \ie{}, $z = [L_i, H_i]$ and $c=\varnothing$.
The diffusion model $\mathcal{G}_i$ learns to denoise $z_{t}$ using the following training loss:
\begin{equation}
\mathcal{L}_{\text{diff}} =\mathbb{E}_{z, c, \epsilon, t} \big[ \left\| \epsilon - \mathcal{G}_i \left(z_{t}, t, c \right) \right\| _2^2 \big].
\end{equation}

Our latent diffusion models $\{\mathcal{G}_i\}$ build upon 3D UNet backbones.
The patch-wise training strategy also serves as data augmentation, helping our model avoid overfitting and learn shared local structures across scenes. 
Our conditional diffusion modeling can capture both lower-frequency structures as well as higher-frequency detail, due to our decomposition of 3D scenes into complementary geometry and latent feature grids at each resolution level.
Our latent diffusion model can then synthesize arbitrary-sized 3D scene outputs by generating scenes in a patch-by-patch fashion. %

\subsection{Large-Scale Scene Generation}
\label{sec:inference}

At inference time, the generation of large-scale 3D scenes is enabled by our hierarchical latent tree representation (\cref{sec:first_stage}) and the latent diffusion models trained on scene patches (\cref{sec:second_stage}).
The generation process reverses the latent tree decomposition and constructs the hierarchy from coarse to fine levels.
At each resolution level, the scene is synthesized in a patch-wise manner, allowing for arbitrary-sized 3D outputs. 

\myparagraph{Patch-by-Patch Scene Generation.}
To construct a latent tree from scratch, we first use the learned diffusion model $\mathcal{G}_1$ to create a coarse scene structure at the lowest resolution level $i=1$.
That is, starting from random Gaussian noise, we apply the diffusion model $\mathcal{G}_1$ to unconditionally generate both the geometry $L_1^s$ and latent feature $H_1^s$ grids of the scene. We find that using a simple inpainting-based patch-generation algorithm 
~\cite{rombach2021highresolution,kim2023nfldm}, without requiring additional networks, works well  to produce  3D scenes without seams.

Patches of $L_1^s$ and $H_1^s$ are synthesized autoregressively on the ground (\ie{}, $xy$) plane.
We adopt a breadth-first patch ordering scheme. From each known patch $z_0$, we first generate its adjacent patches in the $x$ direction and then the $y$ direction; this is then repeated in a ``dilated'' scene generation process. 
To ensure a smooth transition between patches, we require that each patch to be generated by the diffusion model $\mathcal{G}_1$ has partial overlap with existing patches \cite{Lugmayr_2022_CVPR}.
We use the same partial overlap size for patches at all resolution levels. 
Specifically, we follow LDMs~\cite{rombach2021highresolution} and adopt the autoregressive Stable Inpainting scheme:
\begin{equation}
    \label{eq:patch_by_patch}
	z_{t-1} = m \odot z_{t-1}^\text{known} + (1-m) \odot z_{t-1}^\text{unknown},
\end{equation}
where $z_{t-1}^\text{known}$ is sampled from the existing patch $z_0$, while $z_{t-1}^\text{unknown}$ is sampled from the previous denoising iteration $z_t$. These two are then combined using the inpainting mask $m$ to form the new sample $z_{t-1}$, keeping the known region unchanged.

\myparagraph{Coarse-to-Fine Refinement.}
We then refine the generated coarse scene at higher resolution levels.
At each level $i > 1$, the scene grid $L_i^s$ is first reconstructed from the geometry grid $L_{i-1}^s$ and the latent feature grid $H_{i-1}^s$ at the lower-resolution level $i-1$ using the trained decoder $\mathcal{D}_i$, as defined in \cref{eq:latent_tree_decoder}.
Next, the diffusion model $\mathcal{G}_i$ generates the latent feature grid $H_i^s$ conditioned on the geometry grid $L_i^s$ at patch-level with overlap. 

However, generating patches sequentially like Stable Inpainting~\cite{rombach2021highresolution} at higher-resolution levels or infinite scenes with significantly more patches can be extremely time-consuming~\cite{wu2024blockfusion}. To speed up inference, we adapt the denoising fusion scheme from MultiDiffusion~\cite{bar2023multidiffusion}, which takes each denoising step on all patches simultaneously. 
First, we divide the generated coarse 3D scene into $n$ patches with the same overlap size as previous levels. 
Then, the model $\mathcal{G}_i$ performs a denoising step in parallel across all patches from the noisy scene grid $z_t^s$ with geometry patches from $L_i^s$ as the condition. 
Finally, an aggregation step $\mathcal{A}$ blends the patches by averaging the predictions from overlapping regions, ensuring smooth transitions across patches and producing seamless 3D scenes.
Concretely, each denoising step $t$ at scene resolution level $i$ is formulated as:
\begin{align}
    \label{eq:multidiffusion}
    z_{t-1}^j &= \mathcal{G}_i \Big( F_j(z_t^s), t, F_j(L_i^s) \Big), \\
    z_{t-1}^s &= \mathcal{A}\left(\left\{z^j_{t-1}\right\}_{j=1}^n\right),
\end{align}
where $\forall j \in [1, n]$, $F_j$ is an operation that crops the $j$-th patches from the scene grids $z_t^s$ and $L_i^s$, respectively. The latent feature volume $H_i^s$ of the scene is obtained after completing all denoising time steps, \ie{}, $H_i^s = z_0^s$.

The decoders $\mathcal{D}_i$ and diffusion models $\mathcal{G}_i$ are then applied alternately, as illustrated in \cref{fig:pipeline}-bottom right, until the scene geometry $L_N^{s}$ at the highest level $N$ is synthesized as the final output.

\section{Experiments}
\label{sec:experiments}

\begin{figure*}[tp]
	\begin{center}
		\includegraphics[width=1.0\linewidth]{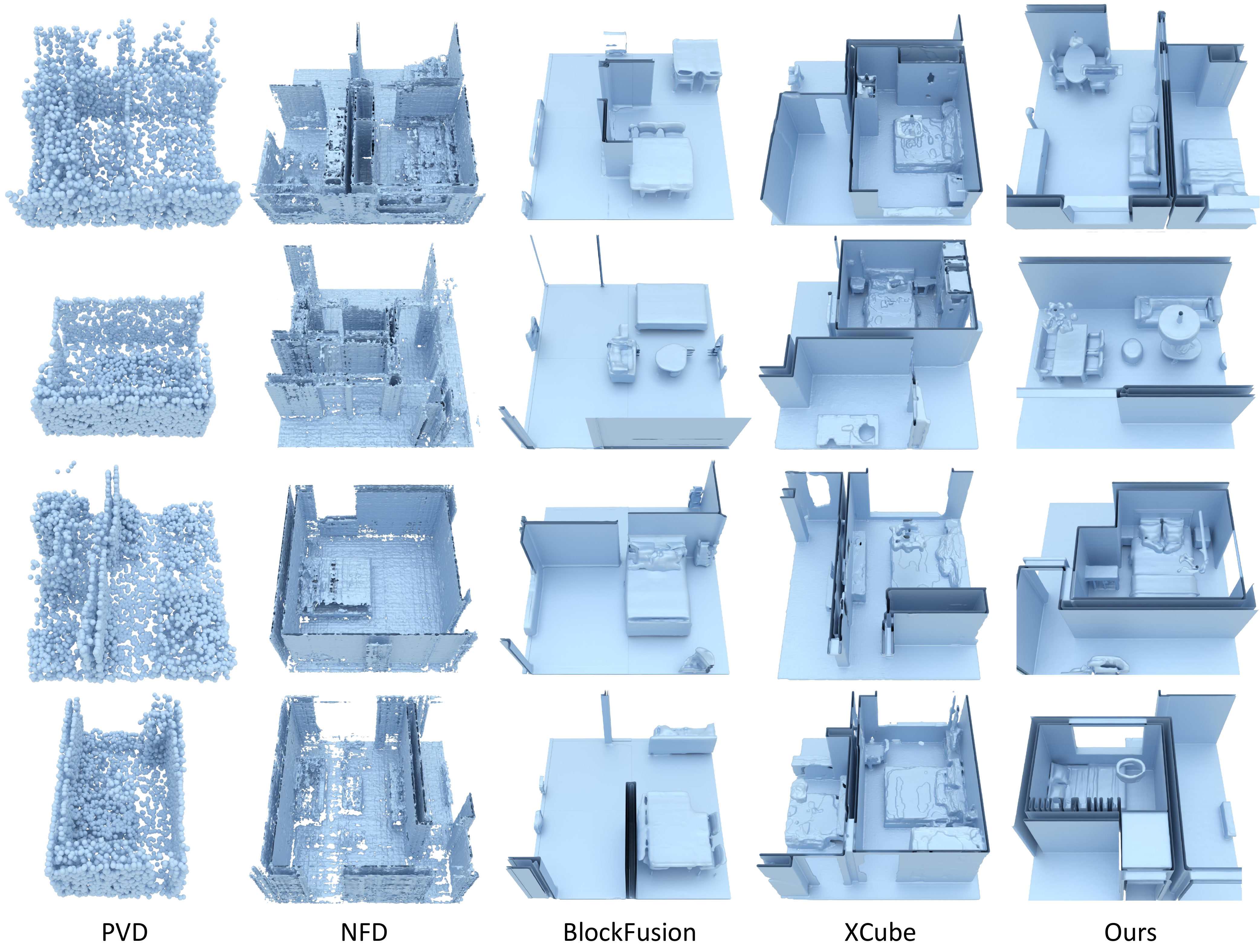}
	\end{center}
	\vspace{-0.7cm}
	\caption{\textbf{Qualitative comparison.} We compare unconditional 3D scene generation with diverse 3D diffusion methods PVD~\cite{Zhou_2021_ICCV}, NFD~\cite{shue20233d}, BlockFusion~\cite{wu2024blockfusion}, and XCube~\cite{ren2023xcube}. All methods were trained on houses from the 3D-FRONT dataset~\cite{fu20213dfront}. Our latent tree-based 3D scene diffusion approach synthesizes cleaner surfaces with more geometric details and captures diverse furniture objects.
 }
	\label{fig:comparison}
 \vspace{-0.5cm}
\end{figure*}

\myparagraph{Experimental Setup.}
We train \OURS{} on house-level scene  data from the 3D-FRONT dataset~\cite{fu20213dfront}, a large-scale indoor scene dataset with diverse structures and layouts. Each house contains various room types (\eg{}, bedrooms, living rooms, kitchens) with different layouts. After filtering out rooms with incorrect furniture scales or no furniture, we retained 6,479 houses. We used an $80\%$/$5\%$/$15\%$ train/validation/test split for both stages of \OURS{} and the baselines. Due to open surfaces in scene meshes, consistent signed distance computation is challenging, so we compute UDFs with a voxel size of $2.2cm$ and a truncation value of $\tau = 10cm$. Instead of preprocessing scenes into fixed patches, we continuously sample chunks from 3D scenes during training to reduce overfitting, applying random flips and rotations for data augmentation.

\myparagraph{Implementation Details.}
We implemented our method in Pytorch and ran all experiments on NVIDIA RTX A6000. 
To first construct latent trees for 3D scenes, we train the encoders and decoders with Adam optimizer~\cite{kingma2014adam}, using a batch size of $4$ and a learning rate of $10^{-4}$, which takes from $5$ hours to one day for different resolution levels until convergence. We use $N=3$ levels in our latent tree. For all levels, the latent feature volumes have $4$ channels. We use the Adam optimizer to train the diffusion models, with a batch size of $8$ and a learning rate of $10^{-4}$, which takes approximately $6$ days until convergence on $2$ GPUs. Note that during the second stage, we infer the latent of each TUDF patch on the fly without storing them on the disk in practice.

\myparagraph{Baselines.}
We compare our method against 3D object diffusion models PVD~\cite{Zhou_2021_ICCV} and NFD~\cite{shue20233d}, as well as the recent 3D scene diffusion approach BlockFusion~\cite{wu2024blockfusion} and XCube~\cite{ren2023xcube}. PVD is based on pointclouds, XCube is based on latent sparse voxels, while 
NFD and BlockFusion are based on triplanes.
All baselines were trained on 3D-FRONT. We uniformly sampled 8,192 points on the 3D scene surface for PVD. We used the same TUDF voxel grids for NFD and XCube as in our approach. The trained model of BlockFusion is provided by its authors, and we followed the authors' instructions for comparisons.

\myparagraph{Evaluation Metrics.}
We follow NFD~\cite{Seitzer2020FID} to evaluate generation ability and use an adapted Fréchet Inception Distance (FID) metric based on rendered images of generated meshes to measure perceptual distribution similarity. We render images of generated and ground-truth 3D meshes by placing cameras at the scene center and rotating them horizontally to capture 20 images per scene. Following prior work~\cite{Zhou_2021_ICCV, erkocc2023hyperdiffusion, ren2023xcube}, we also assess coverage and diversity using the Coverage (COV), Minimum Matching Distance (MMD), and 1-Nearest Neighbor Accuracy (1-NNA) metrics. COV measures the fraction of test set shapes matched to at least one in the generated set, indicating diversity representation. MMD reflects the average distance from each test shape to its nearest neighbor in the generated set, measuring quality and fidelity. The 1-NNA metric evaluates distribution similarity, with a value closer to $50\%$ indicating statistical indistinguishability between the generated and test distributions. We measure point cloud similarity using Chamfer Distance (CD) and Earth Mover’s Distance (EMD). Points are uniformly sampled from the generated meshes of NFD, BlockFusion, our method, and GT, while the same number of points are randomly sampled from PVD’s generated point clouds. Please refer to the supplemental for more details.  

\begin{figure*}[t]
	\begin{center}
		\includegraphics[width=0.95\linewidth]{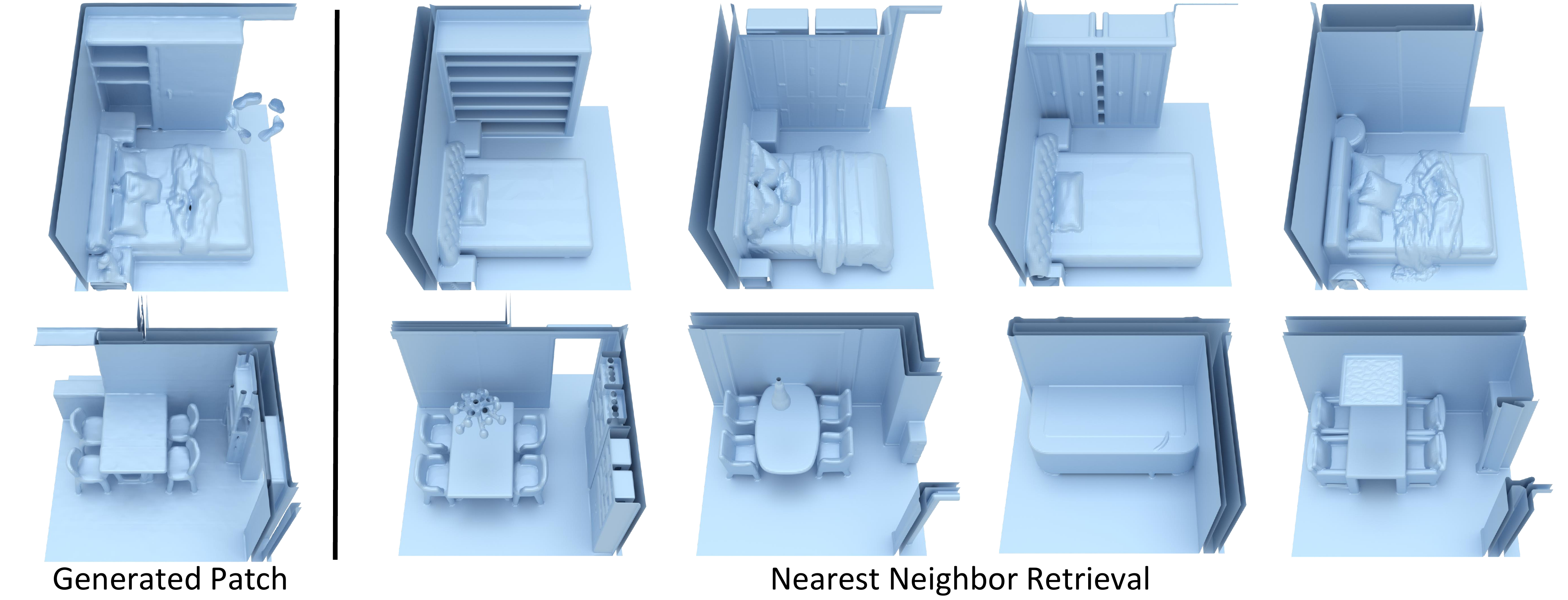}
	\end{center}
	\vspace{-0.6cm}
	\caption{{\bf Generation novelty analysis.} 
     Our generated scene patches (left), compared with their nearest-neighbor retrieved training patches by Chamfer distance. Our approach can synthesize novel patches with different geometric structures than their training set neighbors.
     }
	\label{fig:novelty_analysis}
\end{figure*}

\begin{figure*}[t]
    \vspace{-0.2cm}
	\begin{center}
		\includegraphics[width=0.95\linewidth]{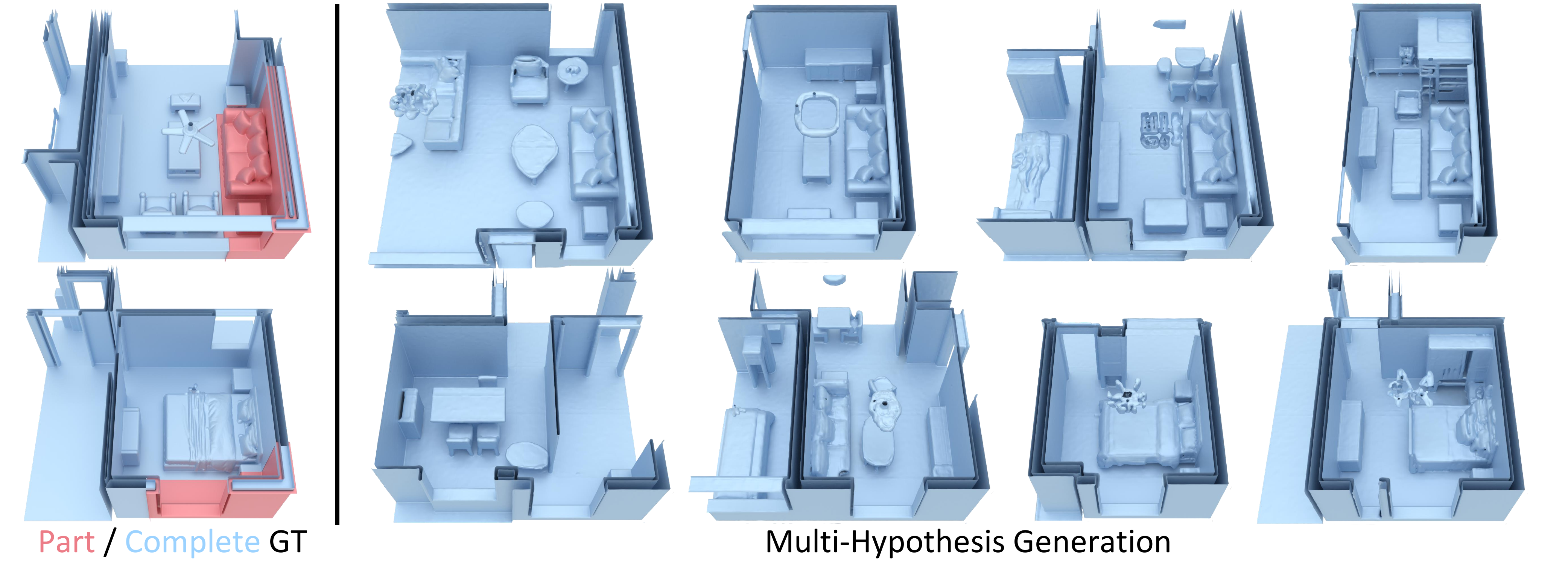}
	\end{center}
	\vspace{-0.6cm}
	\caption{{\bf Probabilistic Scene Completion.} 
     Given an \textcolor{coralpink}{initial part} of a \textcolor{lightblue}{complete ground-truth} 3D scene from the test split, \OURS{} can synthesize various plausible hypotheses of a complete scene by sampling various diffusion outputs.}
	\label{fig:part_completion}
    \vspace{-0.4cm}
\end{figure*}

\subsection{3D Scene Generation}
We evaluate our method on 3D scene generation, both quantitatively and qualitatively in comparison with state-of-the-art 3D diffusion alternatives. We generate 3D scenes of a fixed size for each method for evaluation. 

We show qualitative comparisons in \cref{fig:comparison}. Our results demonstrate superior surface quality and object detail: PVD generates plausible global structures but produces incomplete geometric information in point clouds; NFD~\cite{Seitzer2020FID} generates plausible scene structures and scales well for complex 3D scenes, but struggles to produce complex arrangements of furniture and smooth surfaces; BlockFusion~\cite{wu2024blockfusion} achieves smooth local surfaces and structural diversity, but struggles in generating meaningful higher-level structures (e.g., \cref{fig:comparison}-1st row, where walls are misaligned despite improved surface quality). XCube~\cite{ren2023xcube} produces high-quality scene structures but struggles with local details, like wrinkles on the bed
In contrast, our approach generates scenes with both plausible room structures and layouts while maintaining the highest quality in finer-scale details. 

\begin{table}[t]
    \centering
    \caption{{\bf Quantitative comparison.} We compare unconditional 3D scene generation with state-of-the-art 3D diffusion methods PVD~\cite{Zhou_2021_ICCV}, NFD~\cite{shue20233d}, and BlockFusion~\cite{wu2024blockfusion} trained on 3D-FRONT data~\cite{fu20213dfront}. Our method significantly outperforms baselines in all metrics.
    }
    \vspace{-0.2cm}
    \label{tab:quanComparison}
    \setlength{\tabcolsep}{4.0pt}
    \resizebox{\columnwidth}{!}{
    \begin{tabular}{l|cccccc|c}
        \toprule[1pt]
        \multirow{2}{*}{\centering Method} & \multicolumn{2}{c}{COV (\%) $\uparrow$} & \multicolumn{2}{c}{MMD ($\times 10^{-2}$)$\downarrow$} & \multicolumn{2}{c|}{1-NNA (\%) $\downarrow$} & \multirow{2}{*}{\centering FID $\downarrow$} \\
         & CD & EMD &  CD & EMD  & CD & EMD &  \\ \hline
        PVD~\cite{Zhou_2021_ICCV} & 43.82 & 32.52 & 3.69 & 29.36 & 70.83 & 82.24 & 237.85  \\ \hline
        NFD~\cite{shue20233d} & 44.65 & 45.48 & 3.65 & 27.98 & 62.86 & 63.62 & 266.27  \\ \hline
        BlockFusion~\cite{wu2024blockfusion} & 24.32 & 22.85 & 5.10 & 31.82 & 89.01 & 90.85 & 45.55 \\ \hline
        XCube~\cite{Zhou_2021_ICCV} & 48.60 & 47.71 & \textbf{3.35} & \textbf{26.14} & 56.45 & 59.59 & 55.35 \\ \hline
        \hline
        Ours (17.6 - 2.2) cm & 28.61 & 29.98 & 4.93 & 30.17 & 74.51 & 69.39 & 59.23 \\
        Ours (8.8 - 2.2) cm & 40.87 & 44.21 & 4.90 & 31.50 & 84.34 & 81.57 & 50.07 \\
        Ours (17.6 - 8.8 - 2.2) cm & \textbf{53.10} & \textbf{54.66} & 3.51 & 27.30 & \textbf{53.22} & \textbf{52.56} & \textbf{13.39} \\
        \bottomrule[1pt]
    \end{tabular}}
    \vspace{-0.6cm}
\end{table}

\cref{tab:quanComparison} shows a quantitative comparison of our \OURS{} against baselines. 
Our approach significantly improves both point-based and perceptual-based metrics, demonstrating its effectiveness in 3D scene generation. Since the point-cloud-based PVD~\cite{Zhou_2021_ICCV} generates only sparse point clouds, we convert PVD points to triangle meshes with ball-pivoting~\cite{817351} for FID computation.
The triplane-based NFD~\cite{Seitzer2020FID} produces good scene structures, though lacks perceptual quality in local surface smoothness.
As shown in \cref{fig:comparison}, BlockFusion~\cite{wu2024blockfusion} produces recognizable furniture and smooth surfaces, but with quite fragmented room structures (\eg{}, in \cref{fig:comparison}, the walls in the 1st and 3rd rows are fragmented and unlikely in structure), resulting in a good FID score but a relatively lower point cloud metrics that rely on more strongly reflecting global structures in the ground-truth distribution. XCube~\cite{ren2023xcube} produces compelling 3D scene structure but lacks accurate details by generating complex 3D scenes in a single step without decomposing them to patches.

In contrast, our method follows a hierarchical approach on our latent tree encoding, starting with a good coarse scene structure (reflected in point-cloud metrics) and progressively adding high-fidelity local detail, as indicated by the FID score. 
Additionally, our method is also much more efficient while achieving superior quality: the large-scale scene generation in \cref{fig:teaser} of size $45.1m \times 90.3m \times 2.8m$ containing $\approx 170$ rooms takes 2 hours on a single GPU, while BlockFusion~\cite{wu2024blockfusion} takes 3 hours for the generation of 7 rooms.

\subsection{Ablation Study}

\begin{table}[t]
    \centering
    \caption{{\bf Ablation over latent representation}. Our latent tree-based representation achieves higher reconstruction accuracy while using less training time and latent storage by factorizing 3D scenes into complementary components. }
    \vspace{-0.2cm}
    \label{tab:ablation}
    \setlength{\tabcolsep}{4.0pt}
    \resizebox{\columnwidth}{!}{
    \begin{tabular}{l|cccc}
        \toprule[1pt]
        Representation & \# Params & Training & Storage & Recon. Error ($\ell_2$)\\ 
        \hline
        Latent CM & 13.3 M & $\times$ 1.00 & $\times$ 1.00 & 4.91 $\times$ $10^{-4}$ \\
        Latent Tree (Ours) & 13.3 M & $\times$ 0.87 & $\times$ 0.80 & 3.20 $\times$ $10^{-4}$ \\
        \bottomrule[1pt]
    \end{tabular}}
    \vspace{-0.4cm}
\end{table}

\myparagraph{Latent Tree Structure.}
We validate the latent tree structure of our method in \cref{tab:quanComparison} for 3D scene generation. In this ablation, we compare our full model with latent trees containing only one level. 
As shown in \cref{tab:quanComparison}-bottom, the multi-level latent tree-based diffusion model outperforms the single-level ones. 
Rather than using a single level of the tree and generating a 3D scene in a single step, our \OURS{} adopts a coarse-to-fine strategy for encoding; the diffusion model only generates the higher-frequency component of 3D scenes in each step and, finally, both lower-frequency and higher-frequency components are gracefully combined together and decoded to a high-quality 3D scene.
This ablation demonstrates that the hierarchical tree structure is essential for effective generative modeling with diffusion for complex 3D scenes. 

\myparagraph{Hierarchical Latent Representation.}
We compare our latent tree with latent Cascaded Models (CMs) (\cref{fig:latent_tree_pipeline}-left), commonly used in 2D image synthesis~\cite{ho2021cascaded} and 3D generation~\cite{liu2023pyramid, ren2023xcube}. Unlike CMs, which independently model redundant information at each level, our latent tree decomposes 3D scenes into coarse geometry and fine latent components, capturing residual high-frequency details. As shown in \cref{tab:ablation}, with the same autoencoder and parameter count, our model trains more efficiently (requiring less time per level) and achieves higher reconstruction accuracy (lower $l_2$ error) on test data. Additionally, its more compact latent representation (fewer latent channels) reduces storage requirements when converting TUDF grids to latent grids. Training time and storage statistics are reported as ratios w.r.t. latent CM; please see the supplemental for more details.

\subsection{Novelty of Generated Scene Patches}

An important consideration in unconditional generation is the ability of trained models to produce novel outputs, including new shapes and scene layouts that differ from the training data.
We retrieve the nearest training patch to each generated scene patch based on Chamfer Distance to evaluate this.
We perform retrieval from the entire training set, applying all permutations of the training augmentations (flip and rotation).
In \cref{fig:novelty_analysis}, we show scene patches generated by our \OURS{}, along with the top-4 nearest neighbor training patches. We observe that our generated patches produce new structures different from the training patches. Our method generates strong differences in local object characteristics (top row) as well as differences in layout arrangements (bottom row). We attribute this partly to the random, continuous patch-wise training strategy employed on the fly during the two training stages and the augmentations.

\subsection{Probabilistic Scene Completion}
We further demonstrate the generative capability of \OURS{} for scene completion from partial observations. 
\cref{fig:part_completion} shows that from small portions of a given scene, our diffusion-based approach enables sampling a diverse set of possible completed scenes that explain the partial geometry input. 
Our approach can preserve the initial scene geometry while generating completed scenes of various sizes.
Although \OURS{} is trained without semantic annotations, our completion results respect the input object semantic structures (\eg{}, given a sofa, generating various living rooms).

\section{Conclusion}
\label{sec:conclusion}
We have presented \OURS, a novel approach for high-quality 3D scene generation through latent scene diffusion. We propose decomposing high-resolution 3D scene voxel grids to compact latent trees for efficient diffusion training in this latent space. 
This strategy enables seamless infinite 3D scene generation in a patch-wise manner. Experimental validation demonstrates the capacity of \OURS{} for large-scale 3D scene generation, significantly outperforming state-of-the-art alternatives for unconditional 3D scene generation. 
We hope our method provides an avenue for future exploration towards the challenging task of 3D scene generation and automated 3D content creation.

\myparagraph{Acknowledgements.}
This work was supported by the ERC Starting Grant SpatialSem (101076253). Matthias Nie{\ss}ner was supported by the ERC Starting Grant Scan2CAD (804724). We would like to thank Xuanchi Ren for helpful suggestions on training XCube.

{
    \small
    \bibliographystyle{ieeenat_fullname}
    \bibliography{main}
}

\twocolumn[{
    \renewcommand\twocolumn[1][]{#1}
    \maketitlesupplementary
    \appendix
    \begin{center}
    \centering
        \includegraphics[width=1.0\linewidth]{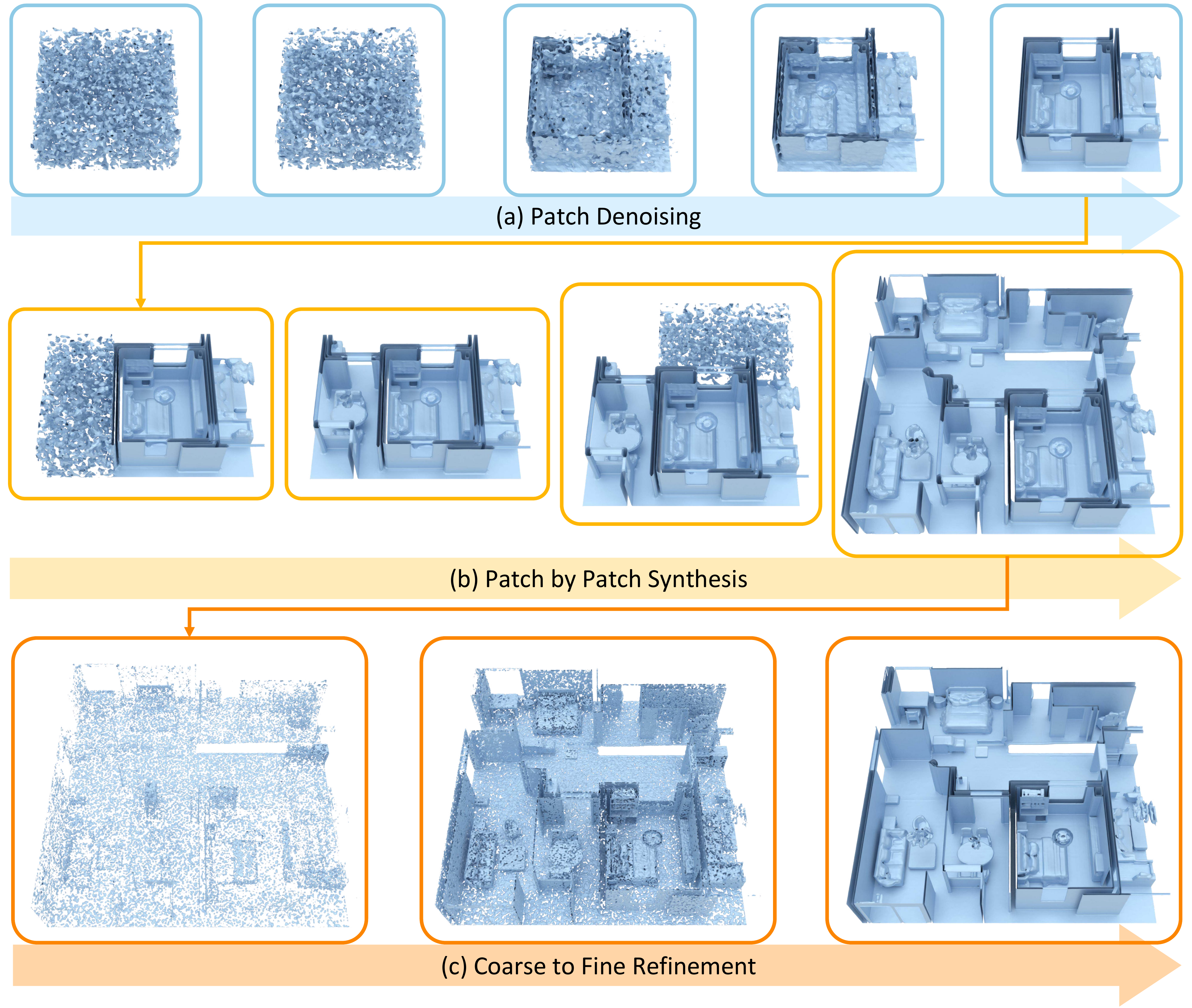}
        \captionof{figure}{\textbf{Intermediate Visualization for 3D Scene Generation.}
         We visualize the DDIM~\cite{song2020denoising} denoising process for (a) patches, (b) the patch-wise inpainting process on the coarse scene as described in \cref{eq:patch_by_patch}, and (c) the coarse-to-fine refinement process as introduced in \cref{eq:multidiffusion}.
        }
    \label{fig:intermediate_vis}
    \end{center}
}]

\begin{figure*}[t]
	\begin{center}
		\includegraphics[width=1.0\linewidth]{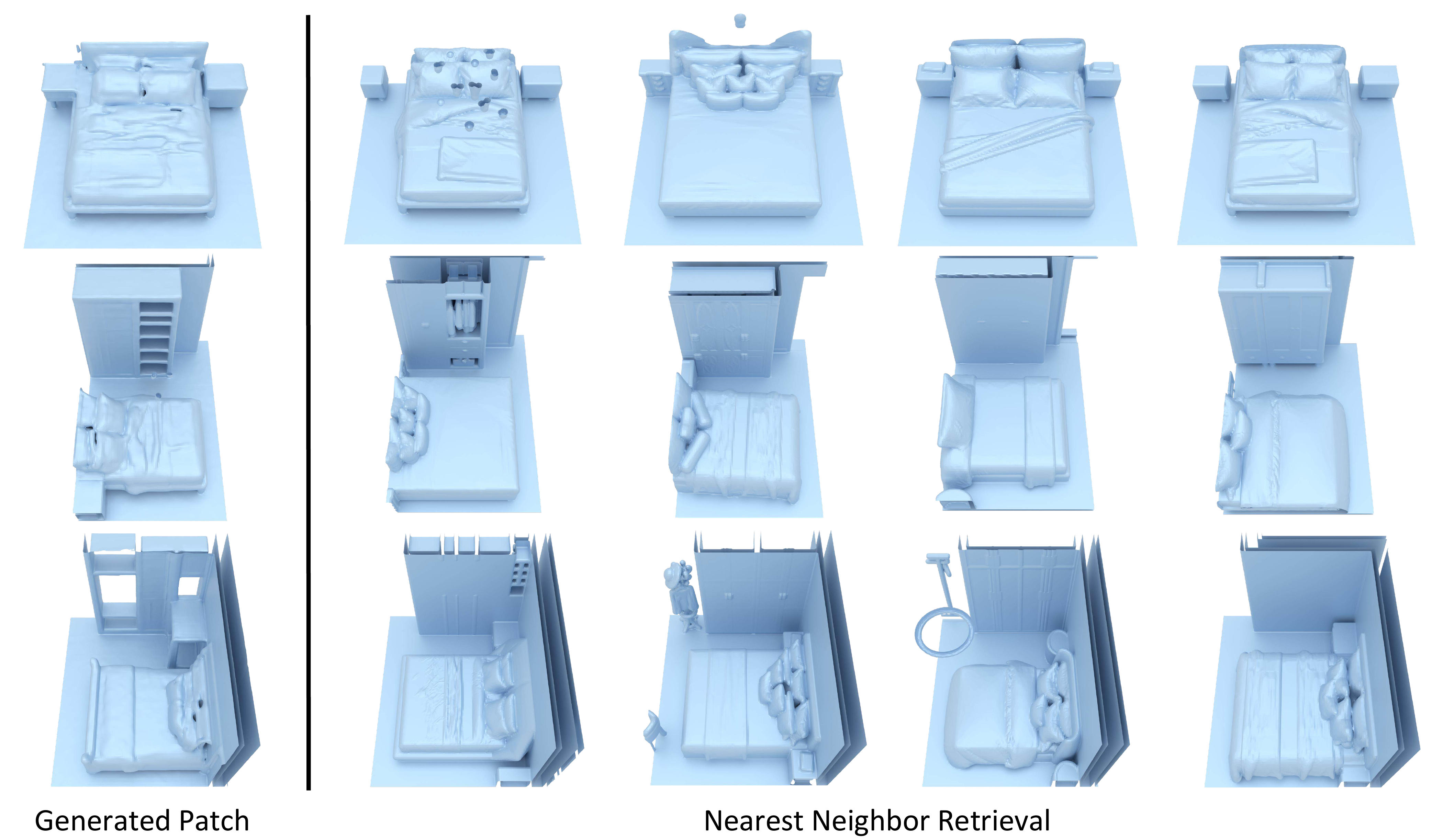}
	\end{center}
	\caption{{\bf Generation novelty analysis.}
     Our generated scene patches (left) are compared with the nearest neighbor's training patches retrieved by Chamfer distance.
    }
	\label{fig:novelty_analysis_supp}
\end{figure*}

\section{Additional Results}
\subsection{Intermediate Visualization}
We visualize the full inference process of generating a 3D scene unconditionally with \OURS{} in \cref{fig:intermediate_vis}. Our method operates patch-wise and coarse-to-fine, as introduced in the main paper. First, starting from the random 3D Gaussian noise  on the left of the first row, \OURS{} gradually denoises it to the mesh at the end of the row. Then, at the second row, our method autoregressively extrapolates the unknown region in a patch-wise manner with an overlap size of one-half of the patch size, which empirically provides sufficient context and reduces seams until the 3D scene of the specified spatial extent is complete. Finally, in the third row, conditioned at the coarse 3D scene, \OURS{} generates the fine details also patch-wise with the same overlap size. This process is implemented batch-wise as each patch is only dependent on the same patch of the previous timestep, \ie{}, all patches at the same timestep are independent of each other. Comparing the coarse-level 3D scene and fine-level 3D scene, we can see that the coarse-level scene determines most of the structure, which ensures a plausible scene layout; the fine-level scene adds rich local detail while keeping the scene structure. The patch-wise and coarse-to-fine 3D scene synthesis enables 3D scene generation with both plausible scene layout and high-quality detail.

\subsection{Additional Infinite 3D Scenes}
Unlike retrieval-based methods (e.g., DiffuScene~\cite{tang2024diffuscene}), focusing on object layouts for single rooms (without walls/structures), our approach synthesizes large-scale scenes, including objects/walls/floors, while maintaining coherent structures at the room scale. 
\OURS{} extends infinitely through outpainting, making it well-suited for open-world games, large-scale robotic training, and film asset creation--a step toward scalable indoor scene generation.

\begin{figure*}[t]
	\begin{center}
		\includegraphics[width=1.0\linewidth]{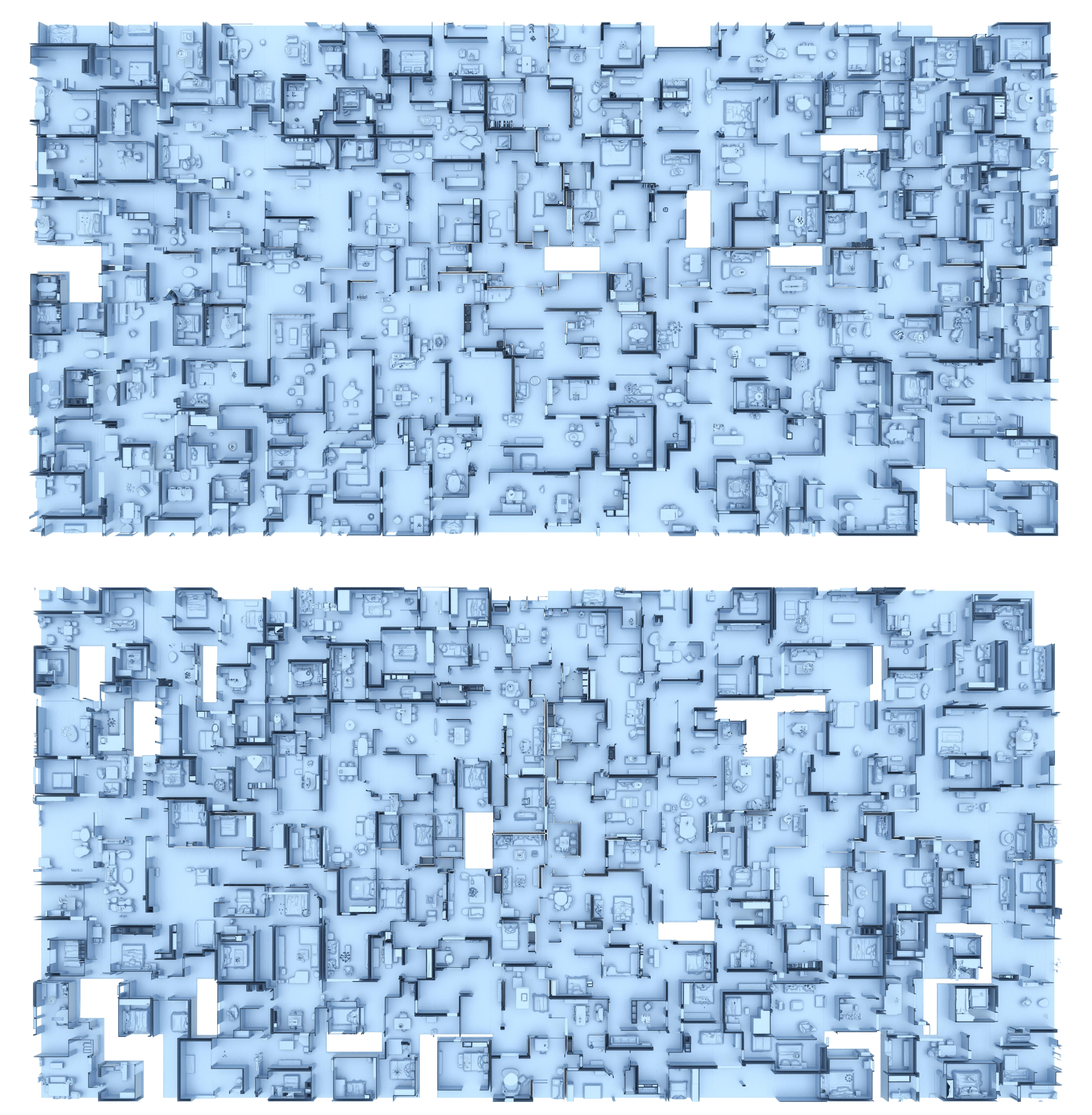}
	\end{center}
	\vspace{-0.2in}
	\caption{{\bf More Infinite 3D Scene Generation Results.}}
	\label{fig:infinite_supp}
\end{figure*}

Here, we present more examples of the infinite 3D scene generation beyond \cref{fig:teaser} in the main paper. In \cref{fig:infinite_supp}, we show two more infinite 3D scenes with the resolution of $[4096, 2048, 128]$ and size of $90.3m \times 45.1m \times 2.8m$. Although trained on the house-level data of 3D-FRONT with only a few connected rooms in each sample, our method generates 3D scenes with diverse room structures, furniture layouts, and varying sizes. 
We generate infinite 3D scenes in the same manner, \ie{}, patch-wise and coarse-to-fine. The supplementary video shows more examples of infinite 3D scenes with zoom-ins and intermediate visualization.

\subsection{Additional Novelty Analysis Results}
In \cref{fig:novelty_analysis_supp}, we provide additional novelty analysis results. %
The results further show that our method learns to generate novel 3D scenes with different furniture layouts and details. We highlight that our model is only trained on random patches of the 3D scenes with complex furniture layouts but without semantic information.

\subsection{Outdoor 3D Scene Generation}
\begin{figure}[t]
	\begin{center}
		\includegraphics[width=1\linewidth]{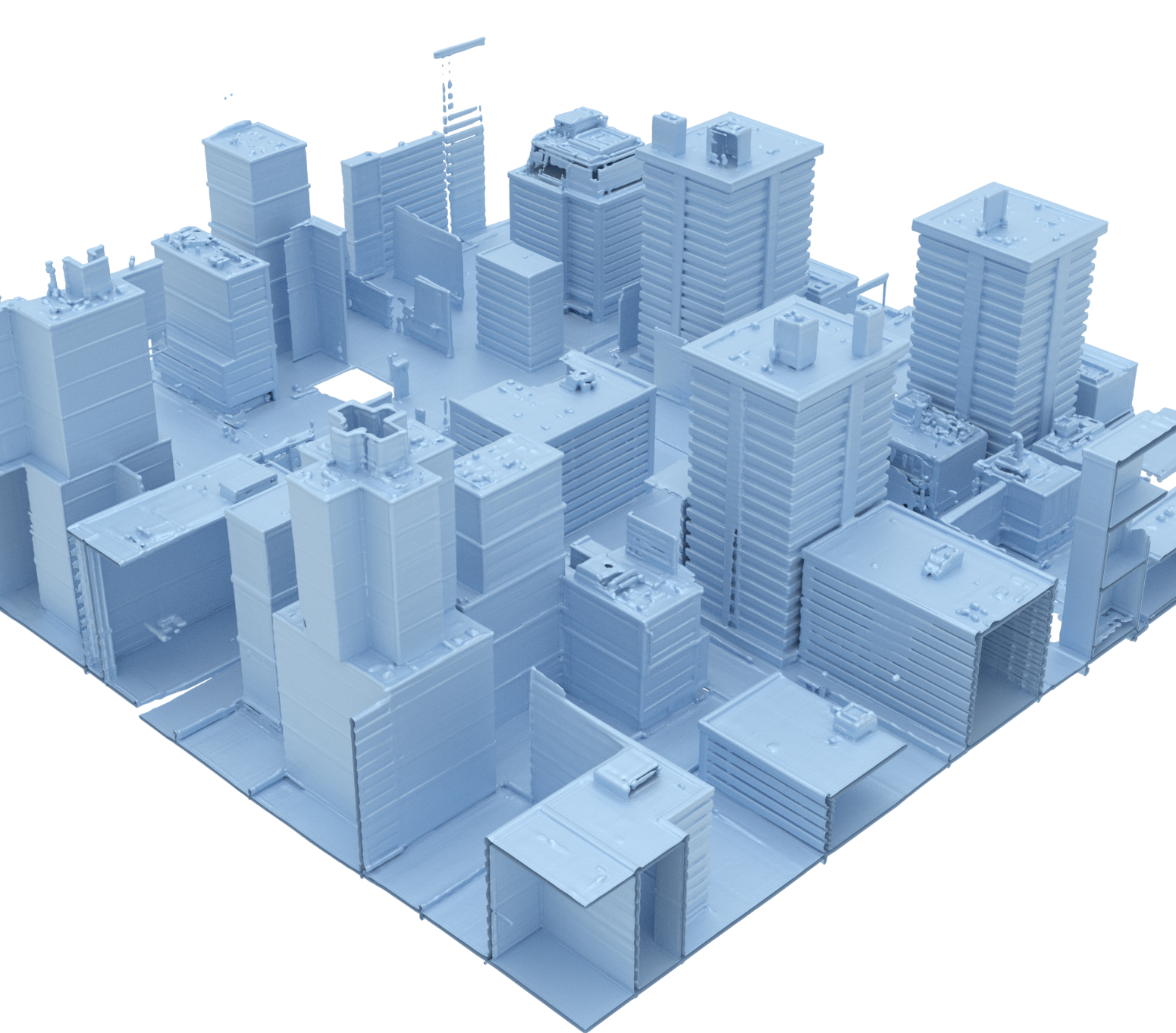}
	\end{center}
	\caption{\textbf{Outdoor 3D Scene Generation.}}
	\label{fig:outdoor_vis}
\end{figure}
In this section, we provide 3D outdoor scene generation visualization in \cref{fig:outdoor_vis}. Unlike methods that assume a specific 3D scene structure, our approach employs a unified voxel-based representation for 3D scenes, allowing seamless extension beyond indoor scenes to diverse environments such as outdoor 3D scenes. As shown in \cref{fig:outdoor_vis}, we train our models on the 3D city asset~\cite{city_rtx2025} in a patch-wise manner, as described in the main paper, and generate large 3D scenes with a resolution of $[1536, 1536, 512]$, where $512$ is the vertical dimension.

\section{Evaluation}
\begin{figure*}[tp]
	\begin{center}
		\includegraphics[width=1.0\linewidth]{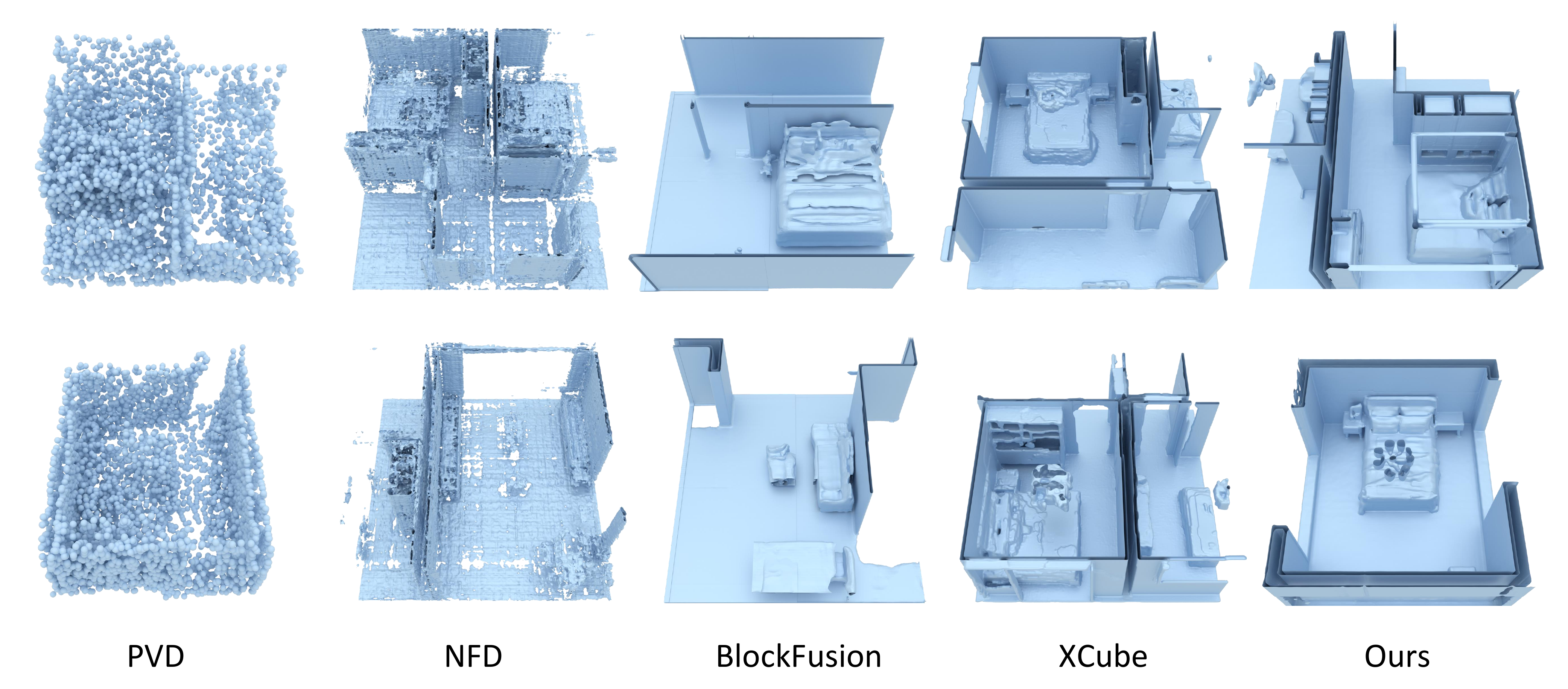}
	\end{center}
	\caption{\textbf{Additional qualitative comparisons.} We compare unconditional 3D scene generation with state-of-art 3D diffusion methods PVD~\cite{Zhou_2021_ICCV}, NFD~\cite{shue20233d}, BlockFusion~\cite{wu2024blockfusion}, and XCube~\cite{ren2023xcube}. All methods were trained on houses from the 3D-FRONT dataset~\cite{fu20213dfront}. 
 Our latent tree-based 3D scene diffusion approach synthesizes cleaner and more detailed surfaces with diverse furniture objects.
 }
	\label{fig:comparison_supp}
\end{figure*}

\subsection{Metrics}
For FID, we set the patch size to be $5.6m \times 5.6m \times  2.8m$ to consider both geometry quality and scene layout. For point clouds-based metrics, we split each generated and ground-truth patch into four smaller ones with the size of $2.8m \times 2.8m \times 2.8m$ to make sure complex structures are represented with a limited number of sampled points. 

The MMD, COV, and 1-NNA metrics are formally defined as: 
\small
\begin{align*}
\text{MMD}(S_g, S_r) &= \frac{1}{\vert S_r \vert} \sum_{Y \in S_r} \min_{X \in S_g} D(X, Y), \\
\text{COV}(S_g, S_r) &= \frac{\left\vert \left\{ \argmin_{Y \in S_r} D(X, Y) \; \middle\vert \; X \in S_g \right\} \right\vert}{\vert S_r \vert}, \\
\text{1-NNA}(S_g, S_r) &= \frac{\sum_{X \in S_g} \Indicator[N_X \in S_g] + \sum_{Y \in S_r} \Indicator[N_Y \in S_r]}{\vert S_g \vert + \vert S_r \vert},
\end{align*}
\normalsize
where $D(\cdot, \cdot)$ represents the CD or EMD distance, $S_g$ and $S_r$ are the sets of generated point clouds and reference point clouds, and $X$ and $Y$ are samples from the generated set and reference set. $\Indicator[.]$ is the indicator function. In the 1-NNA metric, $N_X$ is a point cloud that is closest to $X$ in both generated and reference dataset, \ie{}, 
\begin{equation}
    N_X = \argmin_{K \in S_r \cup S_g} D(X, K).
\end{equation}

\subsection{Baselines}
To train XCube on the FRONT-3D dataset, we follow the open-sourced training scripts in two stages: VAEs and diffusion models. Each stage contains two levels: the coarse level compressing sparse scene patches of shape [128, 128, 64] to dense features of shape [32, 32, 16]; the fine level compressing sparse scene patches of shape [256, 256, 128] to sparse features of shape [128, 128, 64]. The resolution and voxel size of the finest level are determined by the limitations of the GPU memory (NVIDIA RTX A6000) and to remain consistent with other baselines. Since the first two stages only predict voxel occupancy, we utilize the pretrained refinement network of XCube-Objaverse to predict TUDF by making the voxel sizes and the origins of the grids the same. 

\subsection{Additional Comparisons}
We provide additional qualitative comparison results in \cref{fig:comparison_supp}. 

\section{Ablation Study}
\begin{table}[t]
    \centering
    \caption{{\bf Ablation over coarse-to-fine refinement}. Our coarse-to-fine scheme significantly increases the efficiency of large-scale 3D scene generation.}
    \label{tab:ablation_2}
    \setlength{\tabcolsep}{4.0pt}
    \resizebox{\columnwidth}{!}{
    \begin{tabular}{ccc}
        \toprule[1pt]
         & Only Sequential Generation & Full Model (Ours) \\ 
        \hline
        Time & 5 Hours & 2 Hours \\
        \bottomrule[1pt]
    \end{tabular}}
    \vspace{-0.4cm}
\end{table}

\subsection{Hierarchical Latent Representation}
Here, we provide detailed statistics for \cref{tab:ablation} in the main paper: We compare the performance of our latent tree representation against the latent cascaded representation for encoding TUDF voxel grids, using a downsampling/upsampling factor of 4. The latent tree representation requires 20 hours of training and consumes 33KB per room sample for storage, while the latent cascaded representation takes 24 hours and 41KB per sample. After training, we evaluate reconstruction performance on the test set by calculating the $l_2$ error relative to the ground-truth TUDF voxel grids.

\subsection{Coarse-to-Fine Refinement}
For large-scale 3D generation, generating patches autoregressively can be highly time-consuming. In the 3D scene depicted in \cref{fig:comparison}, our full model initially generates the scene patch-by-patch, as outlined in \cref{eq:patch_by_patch}, focusing on capturing the coarse structure. We then apply a coarse-to-fine refinement process, adding high-frequency details using the parallel algorithm, as described in \cref{eq:multidiffusion}. \cref{fig:comparison} shows that our highest resolution synthesis notably improves details in object structures such as chairs and pillows, compared to the coarser resolution shown on the left. In the ablation study (\cref{tab:ablation_2}), we replaced the parallel algorithm with the autoregressive one~\cref{eq:patch_by_patch} to generate the finer levels. The results show that this approach uses $2.5\times$ inference time, compared to our full model. This demonstrates that the batch-wise coarse-to-fine approach significantly reduces overall inference time for large-scale 3D scene generation, in contrast to the purely patch-by-patch method. Furthermore, our approach can be accelerated using multi-GPU setups, a capability that previous works~\cite{wu2024blockfusion, liu2023pyramid}, which employ naive autoregressive patch-wise outpainting, do not support.

\end{document}